\documentclass[letterpaper, 10 pt, conference]{ieeeconf}  

\IEEEoverridecommandlockouts                              

\usepackage{graphics} 
\usepackage{epsfig} 
\usepackage{mathptmx} 
\usepackage{times} 
\usepackage{amsmath} 
\usepackage{amssymb}  

\usepackage{amsmath}
\usepackage{amsfonts}
\DeclareMathOperator*{\argmin}{arg\,min}
\usepackage{graphicx} 
\usepackage{caption}
\usepackage{subcaption}
\usepackage{wrapfig}
\usepackage{algorithm}
\usepackage[noend]{algpseudocode}
\usepackage{booktabs} 
\usepackage{multirow}
\usepackage{color}

\title{\LARGE \bf
   Incorporating Recurrent Reinforcement Learning into Model Predictive Control for Faster Adaption in Autonomous Driving
}

\author{Yuan Zhang$^{1}$, Chuxuan Li$^{2}$, Joschka Boedecker$^{1}$ and Guyue Zhou$^{2}$%
    \thanks{*This work is supported by European Union’s
    Horizon 2020 research and innovation program.}%
    \thanks{$^{1}$Neurorobotics Lab, University of Freiburg, Germany \{yzhang, jboedeck\}@cs.uni-freiburg.de}%
    \thanks{$^{2}$Institute for AI Industry Research (AIR), Tsinghua University, China \{chuxuanli, zhouguyue\}@air.tsinghua.edu.cn}%
}

\begin{document}

\maketitle
\thispagestyle{empty}
\pagestyle{empty}

\begin{abstract}
    Model Predictive Control (MPC) is attracting tremendous attention in the autonomous driving task as a powerful control technique. The success of an MPC controller strongly depends on an accurate internal dynamics model. However, the static parameters, usually learned by system identification, often fail to adapt to both internal and external perturbations in real-world scenarios. In this paper, we firstly (1) reformulate the problem as a Partially Observed Markov Decision Process (POMDP) that absorbs the uncertainties into observations and maintains Markov property into hidden states; and (2) learn a recurrent policy continually adapting the parameters of the dynamics model via Recurrent Reinforcement Learning (RRL) for optimal and adaptive control; and (3) finally evaluate the proposed algorithm (referred as $\textit{MPC-RRL}$) in CARLA simulator and leading to robust behaviours under a wide range of perturbations.  
    
\end{abstract}



\section{INTRODUCTION}
\label{sec:introduction}

    Model Predictive Control (MPC) has become the primary control method for enormous fields, e.g. autonomous driving \cite{reiteremcore} and robotics \cite{song2020learning}. As a model-based method, MPC largely depends on an accurate dynamics model of the system, $x_{k+1} = f(x_k, u_k; \theta)$, where $x, u$ represent state and control respectively and $\theta$ is the parameter of the model. The parameter $\theta$ is assumed to be determined by prior knowledge or system identification method that learns the parameter from a collection of experience. However, the awareness of parameter $\theta$ consistently fails due to the perturbations emerging from all sources in the autonomous driving task. In detail, both internal (e.g. car mass, drag coefficient) and external (e.g. road friction, planning route) parameters may vary in the driving process. Also, the accurate values of the parameters are difficult to collect. Therefore, an MPC controller with a fixed parameter $\theta$ may degenerate the control performance in the autonomous driving task. 
    
    Learning-based MPC is receiving increasing attention as it focuses on automatic adaption to varying environmental parameters~(\cite{hewing2020learningbased,spielberg2021neural}). The standard approach is to incorporate an additional module which aims to tune the parameters of the MPC for optimal control. Bayesian Optimization (BO) is a popular and significant mythology to learn this module. BO aims at optimizing the closed-loop cost $J(\theta)$ and generates the most suitable parameters of an MPC. However, the update of parameters is considerably slow to react in a dynamically changing environment. On the other hand, Reinforcement Learning (RL) is a potential alternative method to searching for the optimal parameters in unknown environments. RL is also capable of modifying the parameters at each time step thus making it a perfect fit for faster adaption. However, the perturbation in the dynamics leads to a non-stationary state variable, thus violating the Markov property required in RL theory. 
    
    This paper follows the RL direction to boost the MPC controller's adaptability. We first reformulate the problem as a Partially Observed Markov Decision Process (POMDP) to ease the non-stationary property under environmental perturbations. The original state in MPC serves as an observation in the POMDP formulation. Meanwhile, a hidden state represents the system's actual state, including the perturbation information. We propose using a recurrent policy to learn such a POMDP with $2$ learning objectives: cumulative reward maximization and system identification loss minimization. The whole system achieves better adaptability in the autonomous driving simulation CARLA compared with pure MPC controller and other variants.



    \begin{figure*}
    \centering
        \begin{subfigure}[b]{0.45\linewidth}
            \centering
            \includegraphics[width=0.8\linewidth]{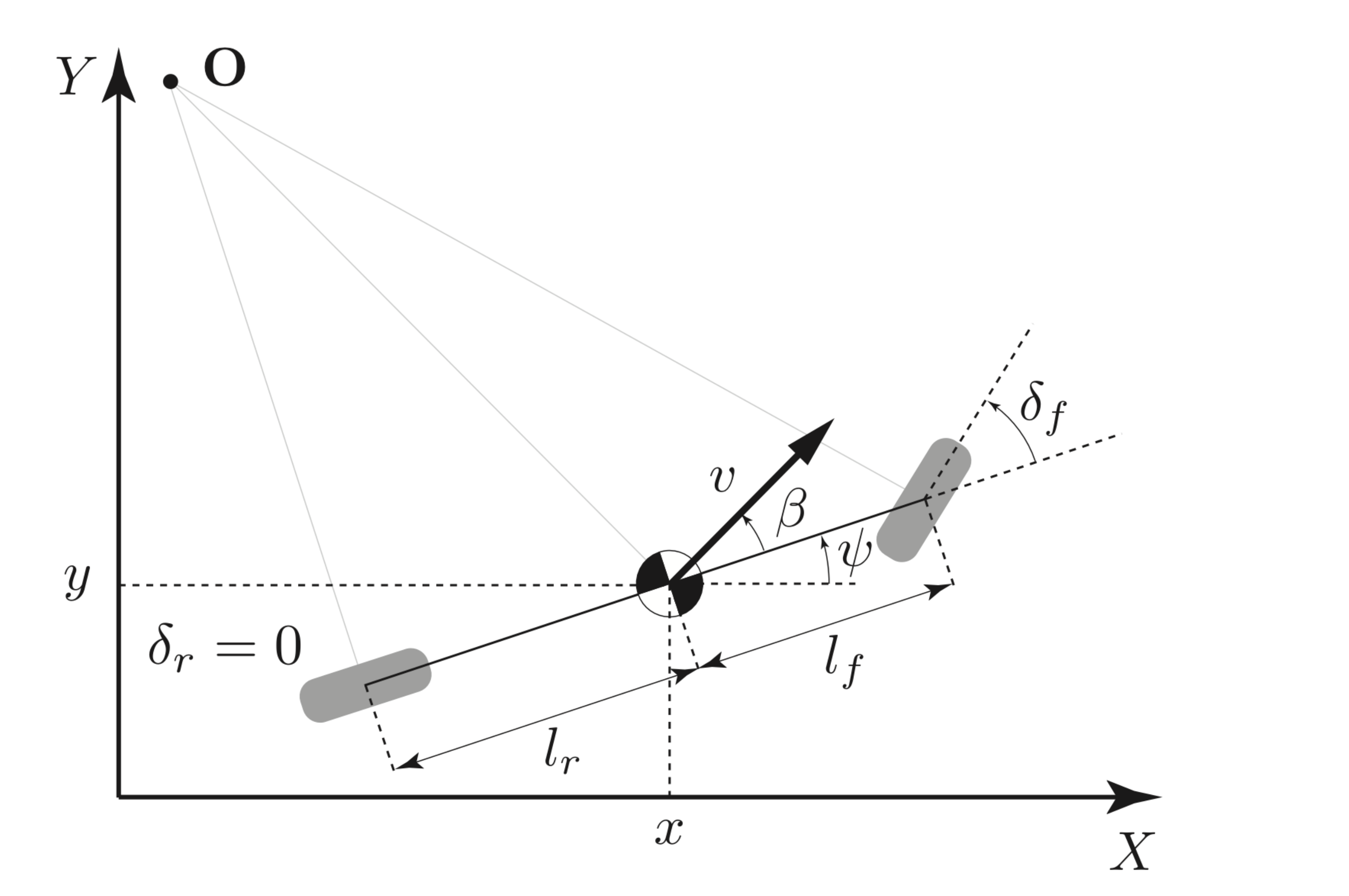}
            \caption{The kinematic bicycle model~\cite{kong2015kinematic}.}
            \label{fig:bicycle}
        \end{subfigure}
        \begin{subfigure}[b]{0.45\linewidth}
            \centering
            \includegraphics[width=0.8\linewidth]{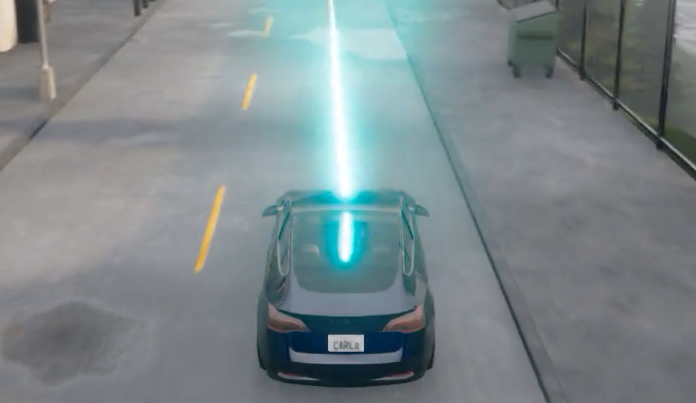}
            \caption{CARLA simulator.}
            \label{fig:carla_simulator}
        \end{subfigure}
     \caption{Model and simulation in the autonomous driving task.}
    \end{figure*}

\section{RELATED WORK}
\label{sec:related_work}

    As mentioned in Section~\ref{sec:introduction}, Bayesian Optimization (BO) is a parallel approach to improve the parameters of an MPC.  \cite{marco2016automatic} adopts an LQR formulation and learns the parameters of $Q$,$R$ matrix with entropy search method. \cite{bansal2017goaldriven} also utilizes an LQR model but learns the parameters of the transition function to control a quadcopter. Both methods update the parameters $\theta$ at the end of an episode as it requires evaluating the cost $J(\theta)$, which leads to slow responses in case of environmental perturbations.

    Comparatively, Reinforcement Learning (RL) can modify MPC parameters at each time step to quickly adapt to dynamic environments.  \cite{zarrouki2021weightsvarying} and \cite{song2020learning} both learn a policy that can improve parameters of MPC's cost function, while \cite{gros2020datadriven} and  \cite{amosdifferentiable} aim to modify both transition and cost function's parameters. Nevertheless, none of these methods consider the non-stationary scenario in an autonomous driving system, which violates the essential Markov property in RL. 
    
    There are other research directions to combine RL with MPC. \cite{brito2021where} utilizes RL to learn a sub-goal so as to reduce the optimization horizon of the MPC. \cite{farshidian2019deep, zhong2013value} replace the terminal cost in MPC by the value function in RL to avoid inaccurate human-designed objectives. While previous works focus on improving the efficiency of MPC with the aid of RL, in our work we are more interested in ensuring adaption to perturbed environments.
    
    
\section{PRELIMINARIES}
\label{sec:preliminaries}

\subsection{Vehicle Dynamics Modelling}
\label{sec:vehilce_dynamics_modelling}

    The kinematic bicycle model~\cite{kong2015kinematic} is a simplified vehicle model targeted for autonomous vehicles, whose continuous time equation is
    
    \begin{equation}
    \label{eq:continuous_bicycle}
        \begin{bmatrix} \dot{p} \\ \dot{q} \\ \dot{\psi} \\ \dot{v} \\ \beta \end{bmatrix} = 
        \begin{bmatrix} v \cos (\psi+\beta) \\ v \sin (\psi+\beta) \\ \frac{v}{l_r} \sin \beta \\ a \\  \tan^{-1} \left( \frac{l_r}{l_f + l_r} \tan \delta_f \right) \\ \end{bmatrix}
    \end{equation}
    
    where state $x=[p, q, \psi, v, \beta]$ includes $p,q$: the coordinates of the mass center, $\psi$: the heading angle of the vehicle, $v$: the speed at the mass center and $\beta$: the angle of the current velocity w.r.t. the longitudinal axis of the vehicle; control $u=[a, \delta_f]$ consists of $a$: the acceleration at the mass center and $\delta_f$: the steering angle of the front wheel. Other than that, $l_r$ and $l_f$ are the vehicle's inertial parameters. An intuitive illustration of the kinematic bicycle model can be seen in Figure~\ref{fig:bicycle}. 
 
    However, such a model cannot be directly adopted in the existing autonomous driving platform (e.g. CARLA~\cite{dosovitskiycarla} and APOLLO~\cite{gao2022apollorl}) where the output control $u$ normally consists of steering $w$, throttle $y$, brake $z$ instead of acceleration $a$ and angle $\delta_f$. \cite{carla_mpc_github} suggests utilizing a neural network to represent the dynamics, which has the following form 
    
    \begin{equation}
    \label{eq:continuous_dynamics} 
        \dot{x} = \begin{bmatrix} \dot{p} \\ \dot{q} \\ \dot{\psi} \\ \dot{v} \\ \dot{\beta} \end{bmatrix} = 
        \begin{bmatrix} v \cos (\psi+\beta) \\ v \sin (\psi+\beta) \\ f_0(v, \beta; \theta) \\ f_1(v, \beta, w, y, z; \theta)  \\  f_2(v, \beta, w, y, z; \theta)  \\ \end{bmatrix} = f(x, u; \theta),
    \end{equation}
    
    where $f_0, f_1, f_2$ are neural networks parameterized with $\theta$, which can be approximated by system identification methods. Furthermore, the discrete dynamic function is simplified written as $x_{t+1} = x_t + f(x_t, u_t; \theta) \Delta t$, with the state $x_t=[p_t, q_t, \psi_t, v_t, \beta_t]^T$ and the control $u_t=[w_t, y_t, z_t]^T$. The control interval $\Delta t$ equals $0.1$ seconds in the experiments.

\subsection{Model Predictive Control Formulation}
\label{sec:model_predictive_control_formulation}

    Model Predictive Control (MPC) is an advanced optimization method for non-linear optimal control problems with constraints. To control an autonomous vehicle moving along a reference trajectory $\textbf{G}=(g_1, g_2, ..., g_{|G|})$ ($g_i$ are $2$-dim coordinated waypoints on the reference trajectory) with a target speed $\textbf{V}$, an MPC problem can be formulated as follows: 
    
    \begin{equation}
    \label{eq:mpc}
    \begin{split}
        \min_{\textbf{x},\textbf{u}} & \quad l_H(x_H, \textbf{G}) + 
        \sum_{t=0}^{H-1} l(x_t, u_t, \textbf{G}, \textbf{V})  \\
        s.t. \quad \forall t,   &\quad x_{t+1} = x_t + f(x_t, u_t; \theta) \Delta t \\
                        & \quad  [-1, 0, 0]^T \preceq u_t \preceq [1, 1, 1]^T \\
                        & \quad\quad x_0 = x_{init}
    \end{split}
    \end{equation}
    
    where $\textbf{x}=(x_0, ..., x_H)$ and   $\textbf{u}=(u_0, ..., u_{H-1})$ represent the state and control sequences to be optimized respectively, and $x_{init}$ is the initial state of the autonomous vehicle. The stage cost function $l(x_t, u_t, \textbf{G}, \textbf{V})= c_{position} * D(x_t, \textbf{G}) + c_{speed} *  |v_t -\textbf{V}| + c_{control} * \|u_t\|_2$, where $D(x_t, \textbf{G}) = \min _{k \in \{1,2,...,|G|\}}  \| (p_t,q_t)^T - g_k\|_2$ is the distance to the nearest waypoint. The terminal cost function $l_H(x_H, \textbf{G}) = \|(p_H,q_H)^T - g_{|G|}\|_2$. Among them, $c_{position}, c_{speed}, c_{control}$ are coefficients to balance different terms in the cost function and set to $0.04, 0.002, 0.0005$ in the experiments. To solve this non-linear MPC problem efficiently, the iLQR method \cite{weiweili2004iterative} can be applied accordingly. 

\subsection{Partially Observed Markov Decision Process (POMDP)}
\label{sec:pomdp}
    \sloppy
    A Partially Observable Markov Decision Process (POMDP) is a generalized mathematical framework of an MDP to deal with the unobserved state issue. It is formulated as a 7-tuple $\langle S, A, P, r, \Xi, O, \gamma \rangle$, where $S$, $A$ and $\Xi$ stand for the state, action and observation space respectively, and $r(s_t,a_t): S \times A \to \mathbb{R}$\ is the reward function at time step $t$. Define $\Delta_{|S|}$, $\Delta_{|A|}$, $\Delta_{|\Xi|}$ be the probability measure on $S$, $A$ and $\Xi$ respectively, then $P(s_{t+1}|s_t,a_t): S \times A \to \Delta_{|S|}$ is the transition function, and the future rewards are discounted by the discount factor $\gamma \in [0,1]$. The most crucial concept in POMDP is that agents can only obtain the observation $o_t$ with probability $O(o_t|s_t, a_{t-1}): S \times A \to \Delta_{|\Xi|}$, instead of receiving the entire state $s_t$.
    
    The received partial observation is not sufficient for agents to make decisions. Instead, the agent needs to maintain a belief state $b_t(s_t): \Delta_{|S|}$ ($b_t$ for short) to estimate a complete knowledge of the system. There exists an update equation for the belief state given previous belief state $b_{t-1}$, action $a_{t-1}$ and new observation $o_t$: $b_t = \eta O(o_t|s_t,a_{t-1}) \sum_{s_{t-1}} P(s_t|s_{t-1},a_{t-1})b_{t-1}$, where $\eta$ is a normalization factor to ensure probability measure. 
    The deterministic policy function in a POMDP is usually defined as $a_t = \pi(b_t, o_t): \Delta_{|S|} \times O \to A$ and the objective of this agent can be formulated as an optimization problem, 
    
    \begin{equation}
    \label{eq:pomdp_obj}
        J^* = \max_{\pi}  \mathbb{E}_{ \pi, P, O} \left[ \sum_{t=0}^{+\infty} \gamma^t r(s_t, a_t)  | b_0 \right].
    \end{equation}
    where $b_0$ is an initial guess on the belief state. The expectation is on policy $\pi$, transition function $P$ and observation function $O$. 

\section{MPC-RRL Framework}
\label{sec:mpc-rrl_framework}

    In this section, we will firstly reformulate the autonomous driving task with an MPC controller as a POMDP problem. Furthermore, within this formulation, we learn a recurrent policy with RL to pursue optimal and adaptive control, thus called \textit{MPC-RRL} for short. 
    
\subsection{POMDP Formulation}
\label{sec:pomdp_formulation}

    As mentioned in Section~\ref{sec:introduction}, the state $x_t$ in an MPC doesn't satisfy the Markov property due to the changing environmental parameters. Instead, we can view $x_t$ as the observation $o_t$ in the POMDP formulation and maintain a hidden state $s_t$ containing the perturbation information in the current system.
    Since the hidden state $s_t$ is unknown, the agent needs to maintain a belief state $b_t$ to get an approximation of the current hidden state $s_t$, as mentioned in Section~\ref{sec:pomdp}. 

    With the conjectural belief state $b_t$, we first design an adaption module $\theta_t = \Omega(b_t)$  that can modify the dynamics' parameters $\theta_t$ at each time step. The MPC controller can utilize the updated parameters $\theta_t$ and current state $x_t$ to calculate the control $u_t$, referred to $u_t = MPC(x_t, \theta_t)$. Thinking the control $u_t$ as the action $a_t$ in the POMDP formulation, the complete policy can be written as $u_t = \pi(b_t, x_t) = MPC(x_t, \theta_t) = MPC(x_t, \Omega(b_t))$. After executing $u_t$ on the autonomous driving system, the hidden state transfers to $s_{t+1}$ with probability $P(s_{t+1}|s_t, u_t)$. The agent can receive the reward signal $r_t=r(s_t, u_t)$ and the next observed state $x_t$ based on the observation distribution $O(x_t|s_t)$, but the exact form of $P$, $O$ and $r$ is unknown to the agent.

\begin{figure*}
        \centering
  	    \includegraphics[width=0.8\textwidth]{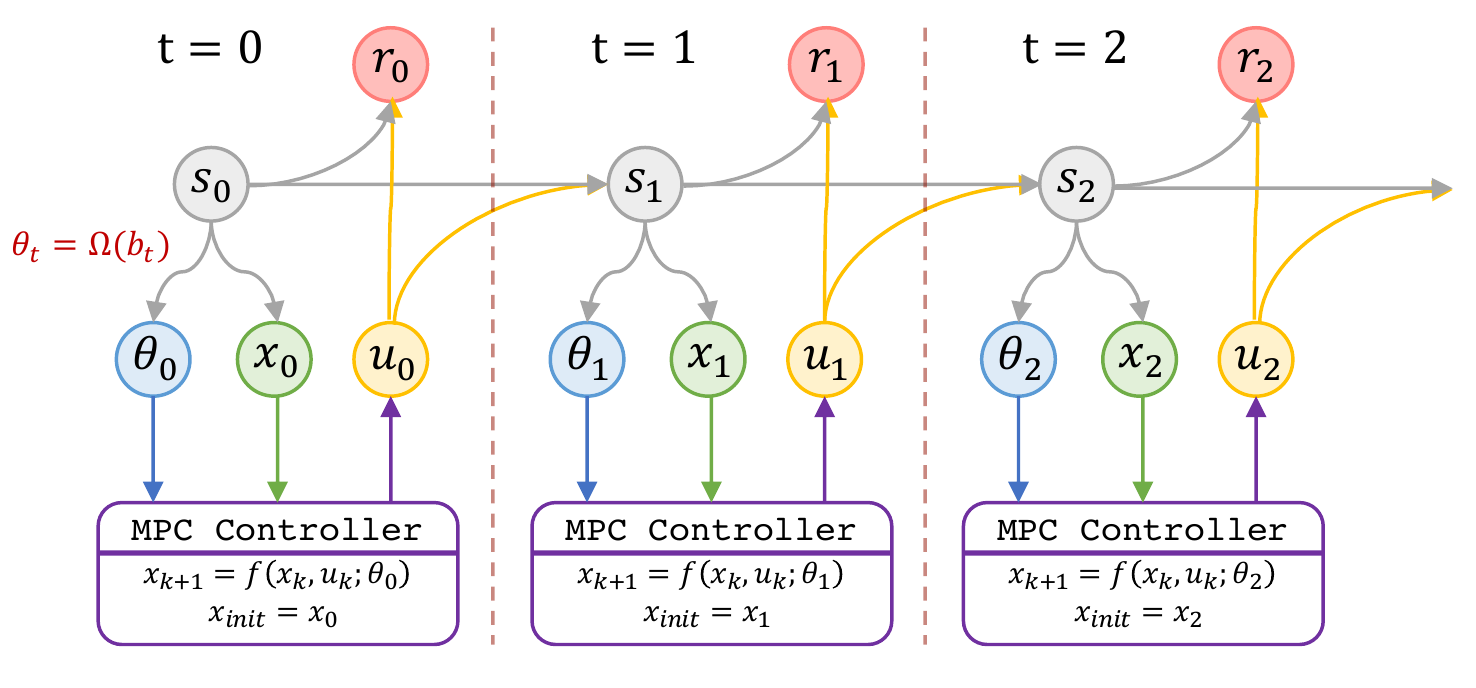}
    \caption{A POMDP formulation of autonomous driving with an MPC controller}
    \label{fig:pomdp}
    \end{figure*}

    This whole POMDP formulation allows the MPC controller to dynamically change the dynamics' parameters, leading to an optimal and adaptive control if the belief state $b_t$ and the action $\theta_t$ is appropriately generated. The overall framework is illustrated in Figure~\ref{fig:pomdp}. 
    
\subsection{Recurrent Policy}
\label{sec:recurrent_policy}
    
    In this section, we will dive deeper into the most essential sub-modules in the framework, the belief state $b_t$ and the policy $u_t = \pi(b_t, x_t)$, on how they are represented, learned and deployed. In general, these two modules are combined in one single recurrent neural network and learned with $2$ objectives for optimality and adaptability respectively. The combination is called recurrent policy for simplicity. 
    
\subsubsection{Policy Representation} 
\label{sec:policy_representation}

    The policy module $u_t = \pi(b_t, x_t)$ consists of an adaption module and an MPC controller. The MPC controller generates control $u_t=MPC(x_t, \theta_t)$ with optimization methods and is free of parameters, thus we only parameterize the adaption module as $\theta_t = \Omega(b_t; \omega)$. As for the belief tracker, we use a deterministic function to represent its update process: $b_{t} = F(b_{t-1}, x_t, u_{t-1}; \tau)$, which is easy to access and optimize. In the end, these two modules with parameters $\omega$ and $\tau$ can have arbitrary structures: neural networks or with prior knowledge.
    
    Following the general practice in POMDP research~(\cite{wang2017learninga,hausknecht2017deepa}), these two modules can be combined into one single recurrent neural network~\cite{hochreiter1997long} (RNN for short). RNN is a type of neural network with memory, which makes it a perfect choice here since the belief state $b_t$ can be seen as the memory of the network and keeps updated. The final combined structure is $b_t, \theta_t = RNN(x_t, u_{t-1}, b_{t-1}; \lambda)$, which perfectly absorbs $F$ and $\Omega$ in one network with parameter $\lambda$. The whole network can thus be optimized and utilized altogether~\cite{young2012pomdpbased}. 

\subsubsection{Execution Details}
\label{sec:execution_details}

    The parameters $\lambda$ are fixed during inference. Given an observation of the vehicle $x_t$, last control $u_{t-1}$ and last belief state $b_{t-1}$ of the system, the agent directly generates the dynamics parameter $\theta_t$ and updates the belief state $b_t$ by a forward pass $b_t, \theta_t = RNN(x_t, u_{t-1}, b_{t-1}; \lambda)$. The MPC controller utilizes the new dynamics parameter $\theta_t$ and initial state $x_t$ to generate the control $u_t=MPC(x_t, \theta_t)$, which is further sent to the autonomous driving environment and get a reward $r_t$ and next state $x_{t+1}$. The specific procedure is illustrated in the Algorithm~\ref{alg:executing_recurrent_policy}. 

    \begin{algorithm}
    \caption{Executing Recurrent Policy}
    \label{alg:executing_recurrent_policy}
    \scalebox{0.85}{
    \begin{minipage}{1.0\linewidth}
    \textbf{Input:} recurrent policy parameters $\lambda$ \\
    \textbf{Output:} trajectory $\tau$ 
    \begin{algorithmic}[1]
    \State Acquire vehicle's initial state $x_1$, initialize vehicle's belief state $b_0$, dummy control $u_0$ and trajectory $\tau=\{\}$
    \For{$t = 1,2,...T+1$} 
        \State Update belief state and parameters with recurrent policy: $b_t, \theta_t = RNN(x_t, u_{t-1}, b_{t-1}; \lambda)$
        \State Calculate vehicle's control with MPC controller: $u_t=MPC(x_t, \theta_t)$
        \State Execute $u_t$ on the autonomous driving system and observe reward $r_t$ and new state $x_{t+1}$
        \State Incorporate experience into the trajectory: $\tau \leftarrow \tau \cup \{u_{t-1}, b_{t-1}, x_t, r_t\}$ 
    \EndFor
    \end{algorithmic}
    \end{minipage}
    }
    \end{algorithm}
    
\subsubsection{Training Details}
\label{sec:training_details}

    The recurrent policy $RNN$ with parameters $\lambda$ is the only module to be learned during training. We design two learning objectives focusing on (i) encouraging MPC to generate optimal control sequences; (ii) adapting the transition model to realities separately. 
    
    The first objective is to maximize the cumulative reward, as the usual objective of POMDP introduced in Equation~\ref{eq:pomdp_obj}, 

    \begin{equation}
    \label{eq:rl_obj}
        J_1 = \max_{\lambda}  \mathbb{E}_{\pi, P, O}\left[ \sum_{t=0}^{+\infty} \gamma^t r(s_t, u_t)  | b_0 \right].
    \end{equation}
    This objective encourages the policy to output the parameters beneficial to the control behaviours with respect to a higher return.

    The second objective is inspired by the system identification loss in control theory, 
    \begin{equation}
    \label{eq:sl_obj}
        J_2 = \min_{\lambda} \mathbb{E}_{\pi, P, O} \left[ \sum_{t=0}^{+\infty} [x_t+f(x_t, u_t; \theta_t)\Delta t  - x_{t+1} ]^2 \right].
    \end{equation}
    This objective pushes the policy to imitate the realistic dynamics' parameters when it deviates from the previous approximations. 
    Combining these two objectives by $J = J_1 + \alpha J_2$ ($\alpha$ is a hyperparameter to balance the objectives), we can successfully learn a policy aiming for optimal and robust MPC control at the same time, which will be shown in the experiment section. 

    In real-world applications, the transition function $P$ and the observation function $O$ is usually unknown to the agents, which makes it difficult to calculate the objective directly. Instead, RL research usually adopts a sampling-based method by learning from the interactions with the environments. Besides, due to the non-differentiable property of the cumulative reward, it's usually suggested to include a value function $Q(x_t, \theta_t; \phi)$ to first approximate the cumulative reward $\sum \gamma^tr_t$ and then learn policies by maximizing this value. In this paper, we follow the same paradigm and conclude the training process in the Algorithm~\ref{alg:training_recurrent_policy}. 
    
    \begin{algorithm}
    \caption{Training Recurrent Policy}
    \label{alg:training_recurrent_policy}
    \scalebox{0.85}{
    \begin{minipage}{1.0\linewidth}
    \textbf{Input:} recurrent policy parameters $\lambda_0$, initial value function parameters $\phi_0$  \\
    \textbf{Output:} final recurrent policy parameters $\lambda_K$ 
    \begin{algorithmic}[1]
    \For{$k =0,1,2,...K$}
        \State Initialize the trajectories set $D_k=\{\}$
        \For{$i =0,1,2,... N$}
            \State Execute Algorithm~\ref{alg:executing_recurrent_policy} to collect a trajectory $\tau_i=(u_0, b_0, x_1, r_1, ...)$ with horizon $T$
            \State Calculate the cumulative reward $R_t=\sum_{j=t}^T \gamma^{j-t}r_j$ for $t=1,...T$ and add them into the trajectory $\tau_i$ 
            \State Enlarge the trajectories set $D_k \leftarrow D_k \cup \{\tau_i\}$
        \EndFor
        \State Update recurrent policy parameters $\lambda$ by minimizing the following objective: \\
            $$\lambda_{k+1} = \argmin_{\lambda} \frac{1}{|D_k|T} \sum_{\tau_i \in D_k}\sum_{t=1}^{T} -Q(x_t, \theta_t;\phi_k)+ \alpha  [x_t + f(x_t, u_t; \theta_t)\Delta t  - x_{t+1} ]^2  $$
            \\ $\quad\quad$  where $b_t, \theta_t = RNN(x_t, u_{t-1}, b_{t-1}; \lambda)$
        \State Update value function parameters $\phi$ by fitting the cumulative reward: 
            $$\phi_{k+1} = \argmin_{\phi} \frac{1}{|D_k|T} \sum_{\tau_i \in D_k}\sum_{t=1}^{T} [R_t - Q(x_t, \theta_t;\phi)]^2  $$
            
    \EndFor
    \end{algorithmic}
    \end{minipage}
    }
    \end{algorithm}

\section{Experiments}
\label{sec:experiments}
    In this section, we evaluate our proposed \textit{MPC-RRL} framework on the CARLA autonomous driving simulator. We first show the superior performance with respect to goal error and route error in the main results and further analyze the controller's behaviours in the ablation study and the policy study. 

\subsection{Experimental Setup}
\label{sec:experimental_setup}

    \begin{figure*}[htbp]
        \centering
  	    \includegraphics[width=1.0\textwidth]{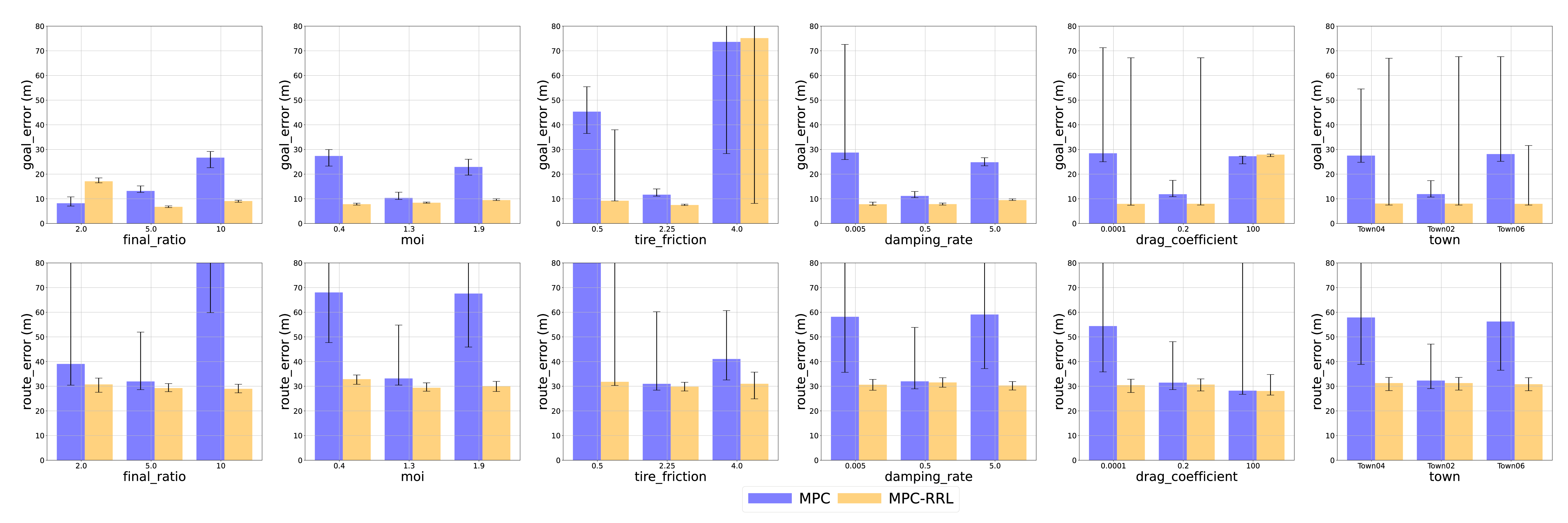}
        \caption{The average goal error and route error of the autonomous vehicle under perturbed testing environments. The $x$-axis is the modified parameter value during testing. The $y$-axis shows the corresponding goal error and route error respectively.}
    \label{fig:main_results}
    \end{figure*}
    
\subsubsection{CARLA Simulator}
\label{sec:carla_simulator}
    We utilize CARLA simulator \cite{dosovitskiycarla} (See Figure~\ref{fig:carla_simulator} for a rendering example) to evaluate autonomous driving performances of all baselines. CARLA simulator is a popular simulation platform to train and evaluate different components of an autonomous driving system (perception, planning, control, etc.). CARLA is grounded on Unreal Engine to run the simulation with changeable configurations. Thus it is convenient to modify both internal (car mass, tire friction) and external (planning route, road friction) factors of the system, which facilitates our evaluation. We run CARLA in a synchronous mode, which ensures the reproducibility of the experiments but may fail to simulate some real-world situations, e.g. missing observations and delayed control, which will be studied in further research.  

 \subsubsection{Task Setup} 
    \label{sec:task_setup}
        
        For each episode of the autonomous driving task, a random starting and goal point is generated. The aim of the control method is to reach the goal point following a reference trajectory. The episode ends if the vehicle reaches the goal or experiences a collision. The performance is evaluated by the goal error and the route error together. The goal error is defined by the distance between the vehicle's final position and the goal point, while the route error is calculated by the cumulative displacement between the vehicle's actual trajectory and the reference trajectory during the driving process. 
        
        We divide the self-driving task into the training and testing phase. We first train the recurrent policy on the default setup of the CARLA simulator and then freeze the policy's parameters afterwards. To evaluate the adaptability of each controller, we modify a system parameter (e.g. car's final ratio, tire's friction) to a different value in the testing phase. A decent controller should manage to adapt to such model mismatches between training and testing environments. The modified parameters and their values during training and testing phase are in Table~\ref{tab:parameter}. Notably, for each parameter, we testify on $3$ perturbed values, with one value slightly different from the training value and the other two deviating largely.
    
    \begin{table}[ht]
    	\caption{Perturbed parameters and their values during training and testing phase.}
    	\begin{center}
    	 \scalebox{0.65}{
    	\begin{tabular}{llll}
		\toprule
			 \textbf{Parameter} & \textbf{Training Value} & \textbf{Testing Value} & \textbf{Explanation} \\
		\midrule
		    final\_ratio & 4.0 & 2.0, 5.0, 10.0 & Transmission ratio from engine to wheels \\
			moi & 1.0 & 0.4, 1.3, 1.9 & Moment of inertia of the vehicle \\
			tire\_friction & 3.5 & 0.5, 2.25, 4.0 & Friction factor of all wheels \\
			damping\_rate & 0.25 & 5e-3, 5e-1, 5e1 & Damping rate of all wheels \\
			drag\_coefficient & 0.15 & 1e-4, 2e-1, 100 & Drag coefficient of the vehicle's car body \\
			town & Town01 & Town04, Town02, Town06 & Carla's default town maps \\
		\bottomrule
		\end{tabular}
		}
    	\end{center}
    	\label{tab:parameter}
    \end{table}

\subsubsection{Baselines}
\label{sec:baselines} 

    \paragraph{MPC Controller} 
    \label{para:mpc_controller}
    \textit{MPC Controller} strictly follows the MPC formulation as described in Section~\ref{sec:model_predictive_control_formulation}. Some practical implementation details should be considered to successfully apply MPC on the CARLA simulator, which will be further clarified in Section~\ref{sec:practical_implementation_details}. 
    
    \paragraph{MPC-RRL Controller}
    \label{para:mpc-rrl_controller}
    \textit{MPC-RRL Controller} is precisely the framework introduced in Section~\ref{sec:mpc-rrl_framework}. For a fair comparison, we adopt the same MPC structure as Paragraph~\ref{para:mpc_controller} as the based controller in the framework but train a recurrent policy for better adaption. 

    \paragraph{MPC-RRL Controller variants}
    \label{para:mpc-rrl_controller_variants}
    We further develop different variants of \textit{MPC-RRL Controller} to evaluate the functions of different designs of the proposed controller. In detail, we respectively exclude one of the three essential designs: RNN structure, system identification loss and cumulative reward and evaluate the variant controller. The results can be found in Section~\ref{sec:ablation_study}. Notably, these variants should cover a lot of branches of adaptive control methods. In detail, using RNN structure and system identification loss alone is a practical way to implement the online estimation of dynamics parameters~\cite{bonassi2022recurrent}. Training with system identification loss and cumulative reward represents the branches of dual control~\cite{wittenmark1995adaptive}. 
    
    Besides, we also try to include a baseline with an end-to-end RNN controller generating control without MPC formulation. However, in practice, we find this controller fails to learn proper driving behaviours even in the training scenario, which implies the difficulties and inefficiencies of pure RL algorithms in such a complex control task. We therefore exclude this result in the following sections.

\subsection{Practical Implementation Details}
\label{sec:practical_implementation_details}

    In this section, we briefly mention some practical implementation details during the experiments, in order to better incorporate the proposed \textit{MPC-RRL} framework into the autonomous driving tasks on the CARLA simulator.

\subsubsection{Vehicle Dynamics Model}

    Regarding the vehicle dynamics model introduced in~\ref{eq:continuous_dynamics}, we adopt $3$ neural networks $f_0, f_1, f_2$ parameterized with $\theta = [\theta_0, \theta_1, \theta_2]$ to represent the unknown parameters. 
    Specifically, $f_0(v, \beta; \theta_0) = vsin{\beta} / \theta_0$ predicts the heading angle $\psi$ in Equation~\ref{eq:continuous_bicycle}. For the formulation of $f_1$ and $f_2$, certain symmetric properties with respect to the velocity angle $\beta$ and the steering control $st$ must be satisfied so that it is aligned with the physical vehicle system: $f_1(v, \beta, w, y, z;\theta_1)= f_1(v, -\beta, -w, y, z;\theta_1)$, $f_2(v, \beta, st,th, br;\theta_2)=-f_2(v, -\beta, -w, y, z;\theta_2)$. To achieve such property, we introduce $2$ auxiliary neural networks $f_1^{'}(v, \beta, st,th, br;\theta_1), f_2^{'}(v, -\beta, -w, y, z;\theta_2)$ which can have arbitrary outputs and further combine these networks to have $f_1(v, \beta, st,th, br;\theta_1)=[f_1^{'}(v, \beta, st,th, br;\theta_1) + f_1^{'}(v, -\beta, -st,th, br;\theta_1)] / 2$ and $f_2(v, \beta, st,th, br;\theta_2)=[f_2^{'}(v, \beta, st,th, br;\theta_2) - f_2^{'}(v, -\beta, -st,th, br;\theta_2)] / 2$ to that $f_1$ and $f_2$ satisfy the symmetric property mentioned before. Specifically, both $f_1^{'}$ and $f_2^{'}$ are $2$-layer feed-forward neural networks with $32$ hidden units for each layer. The activation function is $\tanh$. 
    
    Notably, the prediction of the velocity $v$ should always be positive, for which we further rewrite $f_1(v, \beta, st,th, br;\theta_1) = [f_1^{'}(v, \beta, st,th, br;\theta_1) * (2\sqrt{v} + f_1^{'}(v, \beta, st,th, br;\theta_1)) + f_1^{'}(v, -\beta, -st,th, br;\theta_1) * (2\sqrt{v} + f_1^{'}(v, -\beta, -st,th, br;\theta_1) )] / (2 \Delta t)$ so that $ v + \Delta t * f_1(v, \beta, st,th, br;\theta) \ge 0$ is always true. In general, $\theta=[\theta_1, \theta_2, \theta_3]$ is a set of parameters to describe the dynamics model. 
    
    To provide a warm start for the parameters $\theta$, we further carry out a system identification step by manually controlling the vehicle running in the default map and road setting of the CARLA simulator and collecting $19000$ transitions. The mean-squared-error loss and Adam optimizer \cite{kingma2017adama} are further used to learn the parameters $\theta$. 
    
\subsubsection{Model Predictive Control} 
     As mentioned in Section~\ref{sec:model_predictive_control_formulation}, the distance to the reference trajectory term is written as $D(x_t, \textbf{G}) = \min _{k \in \{1,2,...,|G|\}}  \| (p_t,q_t)^T - g_k\|_2$. However, the minimization operation is difficult in practical gradient-based optimization methods. Instead, we utilize a differential function $D(x_t, \textbf{G}) = -\log \big( \frac{1}{|G|}\sum \limits_{k=1}^{|G|} \exp\{ - \|(p_t,q_t)^T - g_k\|_2\} \big)$ as the distance function, which can be viewed as an approximation and soft version of the original minimization calculation \cite{chen2017constrained}. Other than that, we incorporate the constraints on the controls as a sinusoidal  activation function on the original controls.

\subsubsection{POMDP Setup}
    
    The reward function is a combination of goal error $e_g$ and route error $e_r$, informed as $r_t = \exp\{-e_g / 100.0\} * \exp\{-e_r\} $.
    The state function can be directly acquired from the CARLA simulator. Specifically, position $p$, $q$, heading angle $\psi$ and speed $v$ are read directly, while velocity angle $\beta$ is calculated by $\arctan(v\_y/v\_x)$ and smoothed closed to $0$.
    The action space of the POMDP is $65$-dim, including $1$ parameter in $f_0$, $32$ parameters in $f_1^{'}$ and $f_2^{'}$ respectively (as we only modify the last layer of $f_1^{'}$ and $f_2^{'}$). 

    Regarding the recurrent policy, an LSTM-based neural network~\cite{hochreiter1997long} with a $256$-dim recurrent layer is used. The hidden layer in LSTM can be seen as the belief state in the PODMP framework. To successfully train this recurrent policy, we further adopt an actor-critic algorithm PPO~\cite{schulman2017proximal} as it is one of the best performing on-policy RL algorithms suitable to the training of dynamics hidden states and continuous action space. All network structures and hyperparameters are set equally among all baselines for a fair comparison. The detailed hyperparameters are exhibited in the open-sourced code base due to the page limit.


\subsection{Main Results}
\label{sec:main_results}
    
    We firstly train \textit{MPC-RRL Controller} with environmental parameters at training value until convergence and then include \textit{MPC Controller} into the testing phase. This comparison is fair since \textit{MPC Controller} also identifies the parameters of the dynamics in the same training environment. Each controller is tested for $100$ episodes under each perturbation. From Figure~\ref{fig:main_results}, both MPC-based methods present an acceptable adaptive performance under minor perturbations of environments, which shows MPC do perform robustly against minor perturbations. However, when the perturbed value significantly deviates from the training setup, \textit{MPC Controller} fails to control vehicles towards the goal or follow the reference trajectory while \textit{MPC-RRL Controller} can still complete the task in most cases, except in \textbf{final\_ratio} $2.0$, \textbf{tire\_friction} $4.0$ and \textbf{drag\_coefficient} $100$. It is surprising that \textit{MPC-RRL Controller} performs well even in testing values far from the training ones. We argue that the kinematic bicycle model~\ref{sec:vehilce_dynamics_modelling} is only a simplified approximation to the complex CARLA simulator. And even for the training environment with fixed parameters, the CARLA simulator is not static from the controller's view. As a result, the \textit{MPC-RRL Controller} has to learn a proper abstraction (as the belief state) from the trajectories and generate optimal dynamics parameters for better performance. This in-context information could benefit the fast adaption to unseen parameters during testing.
       
\subsection{Ablation Study}
\label{sec:ablation_study}

   In this section we execute an ablation study on the three most essential designs in the framework: RNN structure (RNN for short), system identification loss (SI for short) and cumulative reward (CR for short). Notably, these variants, which already cover a bunch of adaptive methods, might not have identical structures as previous research but are suitable to this certain autonomous driving task. "- RNN" replaces recurrent policy with a feed-forward policy, which is optimized by $2$ objectives as dual control. "- SI Loss" trains with cumulative reward maximization alone. "- CR" means only minimizing system identification loss, which is a practical way to fit system's parameters online. We follow the same task setup as Section~\ref{sec:main_results} and report the median goal error under all testing values per parameter in Table~\ref{tab:ablation_study}. It indicates that RNN structure plays the most fundamental role in the framework. The combination of these three designs empowers the controller with a more consistent improvement over the adaption ability. 
    
    \begin{table}[ht!]
    \centering
    \caption{The median goal error and average rank of all ablation settings. Less goal error turns to a lower rank.}
    \label{tab:ablation_study}
    \resizebox{1.0\columnwidth}{!}{
    \begin{tabular}{cccccccccc}
    \cmidrule[\heavyrulewidth](r){1-10}
    \multicolumn{3}{c}{\textbf{Ablation Settings}} & \multirow{2}{*}{\shortstack{\textbf{final} \\ \textbf{ratio}}} & \multirow{2}{*}{\textbf{moi}} & \multirow{2}{*}{\shortstack{\textbf{tire} \\ \textbf{friction}}}  &
    \multirow{2}{*}{\shortstack{\textbf{damping} \\ \textbf{rate}}}  & \multirow{2}{*}{\shortstack{\textbf{drag} \\ \textbf{coefficient}}}  &  \multirow{2}{*}{\textbf{town}}& \multirow{2}{*}{\textbf{AVG RANK}} \\
    RNN & SI & CR & & & & & & &  \\    
\cmidrule[\heavyrulewidth](r){1-10}
        + & + & + & 9.09 & 8.44  & 9.13 & 8.33 & 23.50 & 8.03 & \textbf{1.3} \\
        + & + & - & 10.65 & 9.40 & 9.98 & 9.12 & 23.89 & 8.80 & 2.8 \\
        + & - & + & 10.44  & 9.43 & 9.97 & 9.19 & 10.68 & 8.83 & 2.5 \\
        - & + & + & 10.82 & 9.40 & 9.94 & 9.28 & 10.74 & 8.94 & 3.0 \\
\cmidrule[\heavyrulewidth](r){1-10}
    \end{tabular}
    }
\end{table}

\subsection{Study on Recurrent Policy}
\label{sec:study_on_rl_policy}

    \begin{figure}[htbp]
        \centering
        \includegraphics[width=1.0\linewidth]{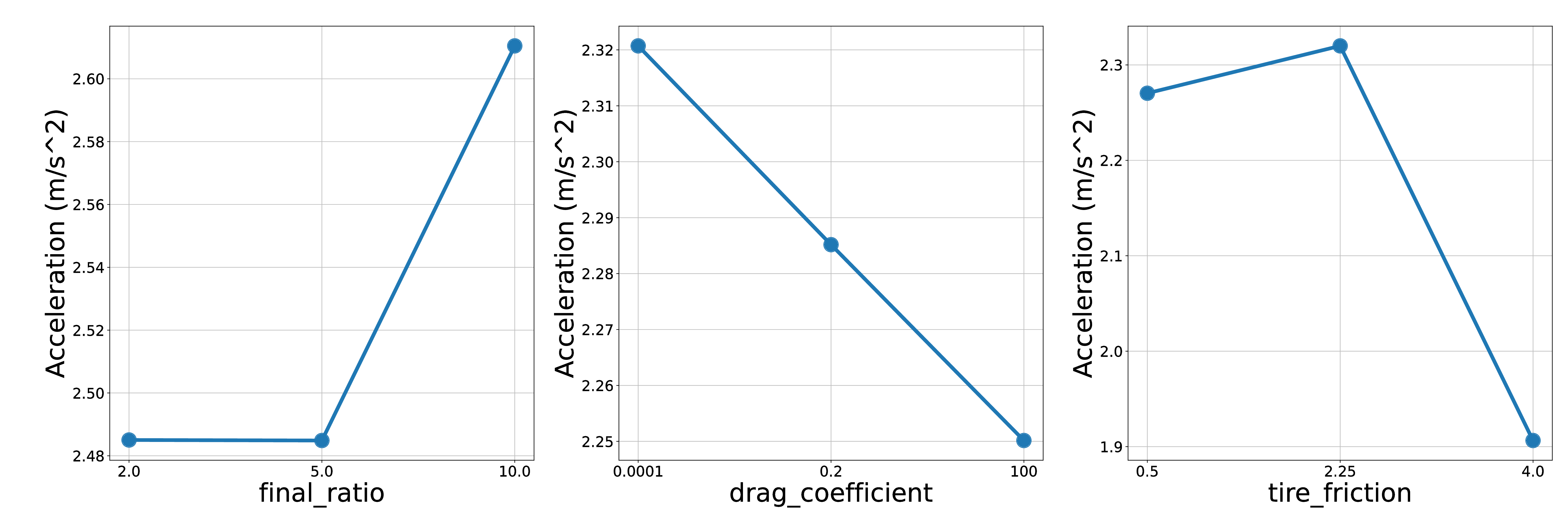}
        \caption{The average absolute value of the autonomous vehicle's acceleration under perturbed testing environments.}
    \label{fig:acceleration}
    \end{figure}

    In this section, we further analyze the learned recurrent policy to find out whether it can really learn the variations of the environmental parameters. We plot a graph on the average absolute value of the vehicle's acceleration and how it varies by the recurrent policy. Figure~\ref{fig:acceleration} clearly indicates that acceleration goes up with the rise of \textbf{final\_ratio} and goes down with the rise of \textbf{drag\_coefficient} and \textbf{tire\_friction}. These trends all show that our \textit{MCP-RRL} framework can adapt well to environmental perturbations, thus leading to better control performance.

\section{CONCLUSION}
\label{sec:conclusion}

    In this paper, we propose an $\textit{MPC-RRL}$ algorithm to handle the problem of perturbed parameters in the autonomous driving task, which is proven to be effective theoretically and empirically. This is the first work to combine RRL and MPC under a POMDP formulation, which could be potentially beneficial to develop more robust control methods. 
    
    In future work, we will combine this work with domain randomization to better generalize the algorithm on unseen environments. Furthermore, $\textit{MPC-RRL}$ is expected to reveal a more robust performance than vanilla MPC on real-world cars, which will be evaluated in further experiments.

\addtolength{\textheight}{-12cm}   




\section*{ACKNOWLEDGMENT}
    This project is funded by the European Union’s Horizon 2020 research and innovation program under the Marie Skłodowska-Curie grant agreement No. 953348. We would also like to thank Jasper Hoffman regarding the usage of SLURM system and Jun Hou for his exploration on CARLA.

\bibliographystyle{IEEEtran}
\bibliography{ref}

\end{document}